%% file: main.tex
\documentclass[lettersize,journal]{IEEEtran}
\usepackage{amsmath,amsfonts}
\usepackage{algorithmic}
\usepackage{algorithm}
\usepackage{array}
\usepackage[caption=false,font=normalsize,labelfont=sf,textfont=sf]{subfig}
\usepackage{textcomp}
\usepackage{stfloats}
\usepackage{url}
\usepackage{verbatim}
\usepackage{graphicx}
\usepackage{cite}
\input{config/packages}
\input{config/formats}

\input{config/pronoun}

\begin{document}

\title{\textit{InverseDraping:} Recovering Sewing Patterns from 3D Garment Surfaces via BoxMesh Bridging}

\author{Leyang Jin\textsuperscript{*}, Zirong Jin\textsuperscript{*}, Zisheng Ye, Haokai Pang, Xiaoguang Han, {Yujian Zheng\textsuperscript{$\dagger$}}, Hao Li

\thanks{L. Jin and Y. Zheng are with the Mohamed bin Zayed University of Artificial Intelligence. Z. Jin, H. Pang, and Z. Ye are with The Chinese University of Hong Kong, Shenzhen. X. Han is with School of Science and Engineering, The Chinese University of Hong Kong, Shenzhen, FNii-Shenzhen, and Guangdong Provincial Key Laboratory of Future Networks of Intelligence. H. Li is with the Mohamed bin Zayed University of Artificial Intelligence and Pinscreen.}
\thanks{L. Jin and Z. Jin contribute equally to this work.}
\thanks{Y. Zheng is the corresponding author. Email: yujian.zheng@mbzuai.ac.ae}

\thanks{Manuscript received 10 October, 2025; revised 25 March, 2026; accepted 26 March, 2026. This work was supported by the Metaverse Center Grant from the MBZUAI Research Office. The work was also supported in part by Guangdong S\&T Programme with Grant No. 2024B0101030002, the Basic Research Project No. HZQB-KCZYZ-2021067 of Hetao Shenzhen-HK S\&T Cooperation Zone, by Guangdong Provincial Outstanding Youth Fund with No. 2023B1515020055, the Shenzhen Outstanding Talents Training Fund 202002, the NSFC with Grant No. 62293482, the Guangdong Research Projects No. 2017ZT07X152 and No. 2019CX01X104, the Guangdong Provincial Key Laboratory of Future Networks of Intelligence (Grant No. 2022B1212010001), and the Shenzhen Key Laboratory of Big Data and Artificial Intelligence (Grant No. SYSPG20241211173853027), the Guangdong Province Radio Science Data Center.}
}



\maketitle
\input{figs/teaser}

\input{sec/0_abstract}

\begin{IEEEkeywords}
Sewing Pattern, 3D Garment Modeling.
\end{IEEEkeywords}

\input{sec/1_intro}

\input{sec/2_related}

\input{sec/4_method}

\input{sec/5_experiments}

\input{sec/6_conclusion}

\bibliographystyle{IEEEtran}
\bibliography{main}

\vfill

\end{document}

%% file: config/packages.tex
\usepackage{graphicx}

\usepackage{booktabs}
\usepackage{orcidlink}

\usepackage{wrapfig}
\usepackage{multirow}
\usepackage{tabularx}
\usepackage[algo2e,ruled]{algorithm2e}

\SetAlFnt{\small}
\SetAlCapFnt{\small}
\SetAlCapNameFnt{\small}
\SetAlCapHSkip{0pt}

\usepackage{mathtools}
\usepackage{balance}

\usepackage{enumitem}
\usepackage{comment}
\usepackage{pifont}

\usepackage{soul}
\usepackage{acronym}
\usepackage{fancybox}
\usepackage{epigraph}
\usepackage{dirtytalk}
\usepackage{bm}
\usepackage{fontawesome}
\usepackage{colortbl} 

\usepackage{xr}
\usepackage{tikz} 
\usepackage{transparent}
\usepackage{subfig}
\usepackage{hyphenat} 
\usepackage{microtype}

\usepackage[capitalize]{cleveref}

%% file: config/formats.tex
\definecolor{cvprblue}{rgb}{0.21,0.49,0.74}
\definecolor{citecolor}{HTML}{0071bc}
\definecolor{frontcolor}{HTML}{325ea5}
\definecolor{backcolor}{HTML}{a58b77}
\definecolor{sidecolor}{HTML}{10768c}
\definecolor{skincolor}{HTML}{dcb7b7}
\definecolor{darkred}{rgb}{0.6, 0.1, 0.05}
\definecolor{DeltaColor}{rgb}{0.039,0.73,0.71}
\definecolor{SigmaColor}{rgb}{0.98,0.45,0.0}
\definecolor{AlphaColor}{rgb}{0,0,0.8}
\definecolor{BetaColor}{rgb}{0.8,0,0.8}
\definecolor{GammaColor}{rgb}{0.514,0.34,0.224}
\definecolor{EpsilonColor}{rgb}{0.353,0.725,0.906}
\definecolor{PurpleColor}{HTML}{9839ff}
\definecolor{BadColor}{HTML}{C0392B}
\definecolor{OrangeColor}{rgb}{0.914,0.541,0.0.141}
\definecolor{GreenColor}{HTML}{00ab41}
\definecolor{LightBlue}{HTML}{7dbaf3}
\definecolor{RedColor}{rgb}{0.949,0.275, 0.224}
\definecolor{LightCyan}{rgb}{0.88,1,1}
\definecolor{Gray}{gray}{0.85}
\definecolor{LightGray}{gray}{0.70}
\definecolor{PinkColor}{HTML}{f37dba}

\definecolor{greenprior}{HTML}{34a853}
\definecolor{redprior}{HTML}{ea4335}
\definecolor{blueprior}{HTML}{4285f4}

\definecolor{bestcolor}{rgb}{1, 0.5, 0.25}
\definecolor{secondbestcolor}{rgb}{1, 0.8, 0.5}

\newcolumntype{a}{>{\columncolor{Gray}}c}

\makeatletter
\newcommand*{\addFileDependency}[1]{%
  \typeout{(#1)}
  \@addtofilelist{#1}
  \IfFileExists{#1}{}{\typeout{No file #1.}}
}
\makeatother

\newlength\savewidth

\crefname{section}{Sec.}{Secs.}
\crefname{table}{Tab.}{Tabs.}
\crefname{figure}{Fig.}{Figs.}
\crefname{equation}{Eq.}{Eqs.}

%% file: config/pronoun.tex
\newcommand{\interRep}{\emph{BoxMesh}\xspace}
\newcommand{\stageI}{\emph{Garment-to-BoxMesh}\xspace}
\newcommand{\stageII}{\emph{BoxMesh-to-Pattern}\xspace}

%% file: figs/teaser.tex

\begin{figure*}[!t]
    \centering
    \includegraphics[width=0.9\textwidth]{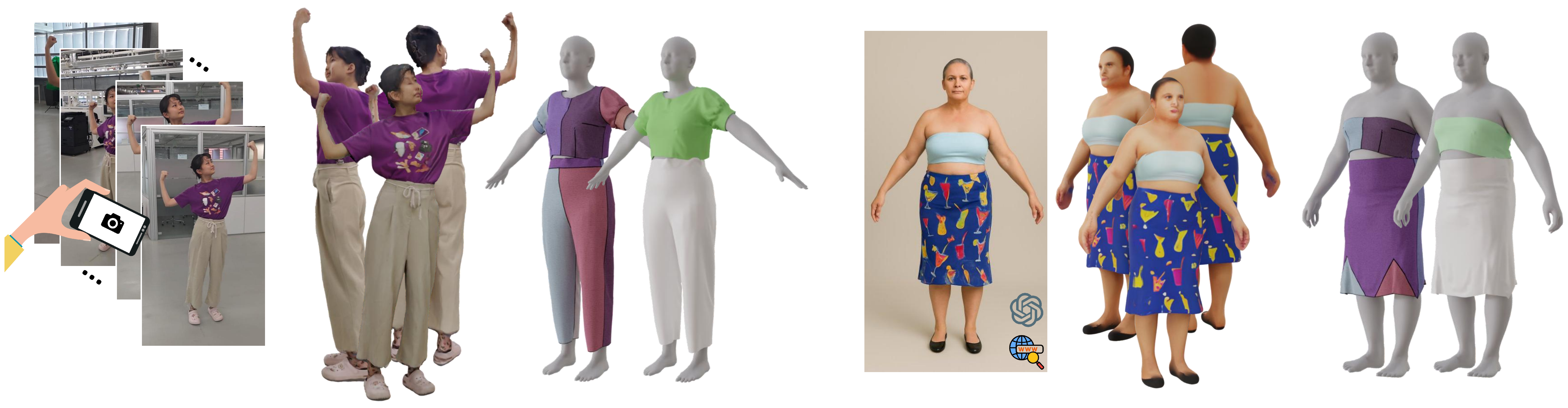}
    \caption{
    Examples of sewing pattern reconstruction from 3D garments obtained via multi-view capture with smartphones (left) and single-view reconstruction (right). Each triplet shows the input, the ``scanned" clothed human from the input, and our predicted sewing pattern.
    }
    \label{fig:teaser}
\end{figure*}

%% file: sec/0_abstract.tex
\begin{abstract}
Recovering sewing patterns from draped 3D garments is a challenging problem in human digitization research. In contrast to the well-studied forward process of draping designed sewing patterns using mature physical simulation engines, the inverse process of recovering parametric 2D patterns from deformed garment geometry remains fundamentally ill-posed for existing methods.
We propose a two-stage framework that centers on a structured intermediate representation, \interRep, which serves as the key to bridging the gap between 3D garment geometry and parametric sewing patterns. \interRep encodes both garment-level geometry and panel-level structure in 3D, while explicitly disentangling intrinsic panel geometry and stitching topology from draping-induced deformations. This representation imposes a physically grounded structure on the problem, significantly reducing ambiguity.
In Stage I, a geometry-driven autoregressive model infers \interRep from the input 3D garment. In Stage II, a semantics-aware autoregressive model parses \interRep into parametric sewing patterns. We adopt autoregressive modeling to naturally handle the variable-length and structured nature of panel configurations and stitching relationships. This decomposition separates geometric inversion from structured pattern inference, leading to more accurate and robust recovery.
Extensive experiments demonstrate that our method achieves state-of-the-art performance on the GarmentCodeData benchmark and generalizes effectively to real-world scans and single-view images.

\end{abstract}

%% file: sec/1_intro.tex
\section{Introduction}
\label{sec:intro}

Sewing patterns are a standard, editable, and highly structured representation of clothing, widely adopted in fashion design and industrial manufacturing. In contemporary garment design workflows, designers typically create sewing patterns using specialized software (e.g., Marvelous Designer, CLO3D, Style3D) and drape them onto digital human models to verify the design. Recently, the inverse problem, recovering sewing patterns from draped 3D garments, has attracted increasing attention due to its applications in virtual try-on, digital wardrobe editing, and reconstructing garments that lack digital design assets. Unlike forward draping, which is well supported by mature physical simulation techniques for mapping flat panels onto 3D bodies, the inverse problem is considerably more challenging, as it requires inferring parametric 2D patterns from deformed 3D garment geometry, a fundamentally ambiguous and underconstrained task.

Prior approaches to inverse garment design are generally limited to simple styles (e.g., T-shirts and pants)\cite{bang2021estimating}, require substantial manual intervention\cite{daanen2008made, decaudin2006virtual, liu20183d, meng2012flexible, wang2003feature, wang2005design, wang2009interactive, yunchu2007prototype}, or rely on predefined templates~\cite{li2024diffavatar}. The most relevant work, NeuralTailor~\cite{korosteleva2022neuraltailor}, attempts to directly learn this mapping using an RNN-based architecture, but shows limited performance on real-world scans. We argue that such end-to-end formulations are inherently ill-posed, as inverse draping involves two coupled challenges: (1) recovering a panel-structured geometric configuration from the draped garment surface, and (2) translating this configuration into parametric sewing patterns. Directly learning this mapping entangles garment structure with draping-induced deformations, making the problem difficult to generalize.

To address this issue, we introduce a structured intermediate representation, \interRep, that bridges 3D garment geometry and sewing pattern parameters. \interRep represents garments as 2D panels arranged in 3D space, with stitching relationships aligned to a rest-pose body. This formulation mirrors the physical garment construction process, where panels are first positioned in 3D before undergoing draping simulation. Crucially, \interRep disentangles intrinsic panel geometry and stitching topology from draping-induced deformations and external effects, yielding a more well-conditioned and physically grounded representation. This formulation decomposes the original ill-posed mapping into two tractable subproblems. \interRep exhibits strong regularities within garment categories while remaining distinctive across types, capturing both global garment-level context and fine-grained panel-level details such as panel shapes, spatial placement, and stitching relationships.

Building on this representation, we propose a two-stage framework for sewing pattern recovery. Given an input garment mesh, we first sample a 3D point cloud and employ a geometry-driven autoregressive model to infer \interRep, reconstructing a panel-structured 3D configuration that captures planar panel geometry and stitching topology. In the second stage, we parse this intermediate representation into parametric sewing patterns using a semantics-aware autoregressive model, which sequentially predicts panel shapes, their 3D placements, and stitching relationships in a structured token space. We adopt autoregressive modeling to naturally handle the variable-length and structured nature of panel configurations and stitching relations. This formulation decomposes inverse draping into geometric reconstruction and structured parameter inference, reducing ambiguity and enabling more accurate and robust recovery.

Extensive experiments demonstrate the effectiveness of our approach. On the GarmentCodeData (GCD)\cite{GarmentCodeData:2024} benchmark, our method achieves state-of-the-art performance. On real-world data, it generalizes well to high-quality human scans from THuman2.0\cite{Zheng2019DeepHuman} and RenderPeople\footnote{\url{https://renderpeople.com/3d-people/}}, as well as to clothed human meshes captured using our cost-effective multi-view setup with commodity devices such as mobile phones. Furthermore, due to its generality, our framework can be combined with methods such as Hunyuan3D~\cite{zhao2025hunyuan3d} to enable single-view reconstruction. In this setting, our method achieves performance comparable to state-of-the-art approaches~\cite{nakayama2024aipparel,bian2024chatgarment}, while in some cases better preserving fidelity to the input image (see \cref{fig:teaser}).

The main contributions of our work are summarized as follows:
\begin{itemize}[leftmargin=*]
\item We show that introducing a structured intermediate representation is key to addressing the ill-posed nature of sewing pattern recovery, and that a staged formulation outperforms end-to-end approaches.
\item We introduce \interRep, which disentangles intrinsic panel geometry and stitching topology from draping-induced deformations, and propose a two-stage framework with autoregressive models tailored for structured prediction.
\item Our method achieves state-of-the-art performance on GCD and generalizes effectively to real-world data, including public scans, smartphone-based multi-view captures, and single-view images.
\end{itemize}

%% file: sec/2_related.tex
\section{Related Work}
\label{sec:related}

\textit{\textbf{3D Garment Digitization.}} The digitization of 3D garments has long been a research focus in computer graphics and vision.
Traditional garment reconstruction approaches primarily utilize techniques like template fitting~\cite{hasler2007reverse,chentemplate}, geometry optimization~\cite{brouet2012design,montes2020computational}, and physics inversion~\cite{inversephys}.
While these methods have demonstrated varying degrees of effectiveness, they are often time-consuming and highly dependent on geometric cues, limiting their ability to generalize across complex garment styles in the real world.
With rapid development in hardware and computational resources, there has been a significant increase in deep learning-based methods enabling garment reconstruction or generation from various types of inputs, including texts~\cite{li2025garmentdreamer}, single-view images~\cite{zhu2022registering, zhu2020deep, lim2024spnet, luo2024garverselodhighfidelity3dgarment, li2024garment, li2025single, sarafianos2025garment3dgen, li2025dress}, multi-view images~\cite{li2024diffavatar}, surround-view videos~\cite{qiu2023rec, zheng2024physavatarlearningphysicsdressed, rong2024gaussian, Pang_2024_CVPR}, 3D point clouds~\cite{ma2021power, zakharkin2021point}, meshes~\cite{Patel_Liao_Pons-Moll_2020,Santesteban_Otaduy_Casas_2019}, and UV maps~\cite{su2022deepcloth}.
However, garments in the real world are frequently coupled with other factors such as human poses and external forces, while many of these approaches consider garment topology as a whole, which hinders their reconstruction quality and limits further applications, including re-editing, simulation, and animation.

\textit{\textbf{Garment Sewing Pattern Representation.}}
Closely aligned with cloth manufacturing processes, sewing pattern depicts all necessary information for the rest state of the garments given a specific body shape, independent of human poses and dynamics. 
Besides mature industrial pipeline creating garments using software (e.g., Marvelous Designer, Style3D), academic interest in sewing patterns has surged over the past few decades.
Recently, synthetic datasets have begun to include sewing patterns as significant components~\cite{wang2018learning, korosteleva2021generating, zou2023cloth4d, liu2023sewformer, GarmentCodeData:2024}, enabling sewing pattern acquisition from 3D point clouds~\cite{korosteleva2022neuraltailor}, single-view images~\cite{liu2023sewformer}, multi-view scans~\cite{li2024diffavatar}, text prompts~\cite{he2024dresscode} and multi-modal inputs~\cite{nakayama2024aipparel, bian2024chatgarment, zhou2025design2garmentcode}.
Additionally, based on the sewing pattern, some methods focus on applications such as garment draping from given pattern~\cite{chen2022structure, li2023isp} and garment reconstruction via optimization with initial template~\cite{li2025dress, li2024diffavatar}.
These advancements highlight the effectiveness of sewing patterns as a structural representation in garment modeling.

\textit{\textbf{Garment Sewing Pattern Acquisition}}
To obtain sewing patterns from garments, earlier methods treat a garment as a combination of developable 3D parts and obtain 2D panels by flattening these pieces.
These methods~\cite{daanen2008made, decaudin2006virtual, liu20183d, meng2012flexible, wang2003feature, wang2005design, wang2009interactive, yunchu2007prototype} not only require ideal conditions of the initial geometries without severe deformation but also involve additional manual control. 
Some more recent studies try to make the procedure automatic by introducing heuristic priors~\cite{bang2021estimating} or learning-based priors~\cite{goto2021data}. 
These methods still heavily rely on garment mesh quality and fail to produce high-quality patterns in more complex cases. 
Recently, ~\cite{korosteleva2022neuraltailor, liu2023sewformer} managed to obtain a sewing pattern by deep neural networks, leveraging data prior from a pattern-level parametric garment dataset~\cite{korosteleva2021generating}.
Unfortunately, these methods tend to overfit to relatively small data scales, preventing their generalization ability to in-the-wild domains like online images or real scans.
Most recent works~\cite{nakayama2024aipparel, bian2024chatgarment, zhou2025design2garmentcode} utilize large language models on the more advanced dataset~\cite{GarmentCodeData:2024}, and achieve appealing results on multi-modal inputs of single-view images and/or text descriptions.
However, the modalities they support only contain rough information, which limits their applications to higher accuracy.

%% file: sec/4_method.tex
\section{Method}
\label{sec:method} 

\input{figs/pattern_rep}

\subsection{Overview}
Our goal is to reconstruct sewing patterns from a given 3D garment $\mathcal{G}$, which is actually an inverse problem of garment draping as shown in~\cref{fig:sewing_pattern} (a). The sewing pattern is well defined in a parametric form.

\textit{\textbf{2D Sewing Pattern Formulation.}}
Following design of ~\cite{GarmentCodeData:2024}, a garment sewing pattern $\{\mathcal{P}, \mathcal{S} \}$ consists of two main elements: a set of $N_P$ panels, denoted as $\mathcal{P}=\{\mathcal{P}_i\}_{i=1}^{N_P}$, and stitching information $\mathcal{S}$. 
Each panel $\mathcal{P}_i$ is represented by a closed 2D polygon, which is defined by a list of $N_i$ edges, $\mathbf{E}_i = \{E_{i,j}\}_{j=1}^{N_i}$. 
Every edge $E_{i,j}$ can be modeled as a straight line, a quadratic or cubic Bézier curve, or an arc. 
We generally represent the shape of edge $E_{i,j}$ by a set of parameters $\mathbf{c_{i,j}}$ as local coordinated control points or the radius and orientation of an arc.
An example can be found in~\cref{fig:sewing_pattern} (b).
Furthermore, each panel $\mathcal{P}_i$ includes a 3D rotation $R_i\in\text{SO}(3)$ and a translation $T_i \in \mathbb{R}^3$, capturing the relative orientation and position of $\mathcal{P}_i$ with respect to the human body. 
These 3D placement parameters $R_i$ and $T_i$ are essential for draping each fabric onto the body model in simulation engines. 
The stitching information $\mathcal{S}$ is defined as pairs of edges $\{E_{i,j}, E_{i', j'}\}$, indicating that the $j$-th edge of panel $\mathcal{P}_i$ is sewn to the $j'$-th edge of panel $\mathcal{P}_{i'}$. \\

However, we find it challenging to obtain the sewing pattern in such representation directly by some end-to-end fully regressive architectures~\cite{korosteleva2022neuraltailor, liu2023sewformer}, since there is a huge gap between the pattern parameters and the 3D shape of the input draped/scanned garment mesh.
To tackle this issue, we simplify this problem into two stages, using a unique \interRep representation as an intermediate bridge. 

\textit{\textbf{BoxMesh Representation.}}
A \interRep of a sewing pattern is a bounding mesh surrounding a human body in a rest pose.
Specifically, it is formed by two parts: 3D panels $\tilde{\mathcal{P}}$ and 3D stitches $\tilde{\mathcal{S}}$. 
When constructing 3D panels, every vertex $\mathbf{x}_k  (1\leq k\leq N_i)$ of each 2D panel $\mathcal{P}_i$ is transformed to point $\mathbf{X}_k$ in 3D space:
\begin{equation}
    \mathbf{X}_k = R_i \tilde{\mathbf{x}_k} + T_i,
\end{equation}
where $R_i$ and $T_i$ are the rotation and translation vector corresponding to panel $\mathcal{P}_i$, and $\tilde{\mathbf{x}_k}$ is padded by $\mathbf{x}_k$ to $\mathbb{R}^3$.
By this way, each 3D panel $\tilde{\mathcal{P}_i}$, $N_i$ 3D edges $\tilde{\mathbf{E}} = \{\tilde{E}_{i,j}\}_{j=1}^{N_i}$ can be calculated similarly, and $N_P$ 3D panels $\tilde{\mathcal{P}}=\{\tilde{\mathcal{P}_i}\}_{i=1}^{N_P}$ can be obtained. 
The stitching parts $\tilde{\mathcal{S}}$ also preserve the same connection rules as 2D edges.
In practice, the \interRep $\mathcal{M}$ can be obtained by mesh triangularization techniques~\cite{eberly2008triangulation}. \\

The visual illustration of the sewing pattern and the draping process can be found in~\cref{fig:sewing_pattern}. 
It is worth mentioning that the \interRep is usually used in cloth simulation from a designed sewing pattern, but what we manage to prove in the opposite direction: extracting sewing patterns from existing 3D garments.
We are the first to leverage the \interRep in the sewing pattern recovery task.

Under the above formulation, we propose a fully automatic two-stage method: inverse simulation and sewing pattern parsing. As illustrated in~\cref{fig:pipeline}, Stage I (\cref{subsec: stage1}) recovers an intermediate representation \interRep $\mathcal{M}$ from the input 3D garment surface $\mathcal{G}$. This effectively implements the inverse simulation in the draping, transforming the garment surface back into a geometric configuration organized by flat panels. The resulting \interRep encodes complete information about panel planes and stitching, but this information is entangled within a 3D mesh rather than expressed as explicit sewing pattern parameters. In Stage II (\cref{subsec: stage2}), we further parse the \interRep $\mathcal{M}$ to recover the sewing pattern parameters ${\mathcal{P}, \mathcal{S}}$.

\input{figs/pipeline}

\subsection{Stage I: Inverse Simulation}
\label{subsec: stage1}
In the first stage, our objective is to predict a \interRep $\mathcal{M}$ from the input 3D garment $\mathcal{G}$.
Following previous work~\cite{korosteleva2022neuraltailor}, we sample a point cloud $\mathbb{P}_{garment}$ from $\mathcal{G}$ as the input, because the current neural encoders of points are relatively mature~\cite{chen2024meshanythingv2,zhao2024michelangelo}.
At its core, our task can be regarded as a mesh generation problem. With the rapid advances in auto-regressive transformers, meshing, i.e., extracting mesh surfaces from point clouds, has recently achieved remarkable results, often reaching artist-created quality, with MeshAnything~\cite{chen2024meshanythingv2} as a representative work. Motivated by this progress, we also explore such approaches for our task.

However, the application is not straightforward. When directly applying MeshAnything~\cite{chen2024meshanythingv2} to our inverse simulation setting, we observe a lack of local details, such as hems, collars and sleeves in the reconstructed \interRep (as shown in the last column of~\cref{fig:eval_stage1}). This arises because, although our task falls under mesh generation, it is not a conventional meshing problem: the input and output 3D shapes are spatially misaligned due to non-rigid deformation. Specifically, a \interRep is composed of multiple planar panels with abundant sharp features, while a draped garment undergoes physical simulation and appears smoother under the influence of gravity and fabric properties. The direct tokenization strategy used in MeshAnything~\cite{chen2024meshanythingv2} directly discretizes coordinates into tokens, which produces overly explicit tokens that preserve excessive 3D spatial information of \interRep, making them unsuitable when input and output are not geometrically aligned.
More importantly, the deformation between the \interRep and the draped garment depends on interactions between neighboring vertices. The direct tokenization strategy enforces a strict bottom-to-top order in token arrangement and does not fully account for local vertex interactions during mesh generation.

Inspired by recent work on BPT~\cite{weng2024scaling}, we adopt Compressive Tokenization instead. Although originally designed to handle meshes with larger face counts, we find it particularly suitable for our task. The compression weakens the strict spatial dependency between input and output shapes, while the incorporated patchified aggregation within the 1-ring neighborhood encourages the model to capture local interactions during mesh generation. As a result, Compressive Tokenization~\cite{weng2024scaling} is better aligned with the requirements of our inverse simulation problem.

According to~\cite{chen2024meshanythingv2}, the garment points $\mathbb{P}_{garment}$ are passed to the autoencoder $\mathcal{E}_1$ of 3D shape generation model~\cite{zhao2024michelangelo} to extract the point cloud feature $\mathbf{F}$.
Then we employ a mesh auto-regressive transformer $\mathcal{O}_1$ to predict \interRep mesh face token sequence $\mathbf{M}^p$ from $\mathbf{F}$.
Finally, the token sequence $\mathbf{M}^p$ is detokenized to \interRep mesh $\mathbf{M}$.
The whole process can be expressed as follows:
\begin{equation}
\tilde{\mathbf{M}} = \mathcal{D}_{seq2boxmesh}(argmax(\mathcal{O}_1(\mathcal{E}_1(\mathbb{P}_{garment})))),
\end{equation}
where $\mathcal{D}_{seq2boxmesh}(\cdot)$ is the detokenizer of Compressive Tokenization mechanism in~\cite{weng2024scaling}.

To prepare training data, we employ Compressive Tokenization to discretize and tokenize $\mathbf{M}$, resulting in a face token sequence $\mathbf{M}^d$. 
$\mathbf{M}^d$ is concatenated with $\mathbf{F}$ as the sequence $Q = [\mathbf{F} | \mathbf{M}^d ]$.
During the training process, we finetune the point cloud feature extractor $\mathcal{E}_1$ and the auto-regressive transformer $\mathcal{O}_1$ using cross-entropy loss on the \( \mathbf{M}^d \) and the predicted face token sequence probabilities \( \mathbf{M}^p \) as follows:
\begin{equation}
\mathbf{M}^p = \mathcal{O}_1([\mathcal{E}_1(\mathbb{P}_{garment})\ |\ \mathbf{M}^d]),
\end{equation}
\begin{equation}
\mathcal{L}_{shape} = - \sum_{i=1}^{M} \sum_{j=1}^{L} \mathbf{1}(\mathbf{M}^d_j = i) \log p(\mathbf{M}^p_j = i),
\end{equation}
where \( M \) is the discretised resolution of $\mathbf{M}$, and $L$ is the length of \( \mathbf{M}^d \) and \( \mathbf{M}^p \).

\subsection{Stage II: Sewing Pattern Parsing}
\label{subsec: stage2}
In the second stage, our objective is to recover parameterized sewing patterns from the predicted \interRep $\mathcal{M}$.
A straightforward solution would be to parse \interRep via mesh segmentation to separate different panel planes and stitching, followed by step-by-step extraction to obtain sewing pattern parameters.
However, such a heuristic pipeline is prone to be error accumulation, which prevents accurate recovery.
Moreover, these approaches encounter significant challenges when applied to the GarmentCodeData (GCD)~\cite{GarmentCodeData:2024}, which contains complex garment styles with up to 37 panels per garment, 37 edges per panel, 104 stitch pairs, and diverse edge types, such as quadratic/cubic Bézier curves, and circular arcs.
Besides, panels usually intersect with each other within a \interRep in 3D space.
Handling such complexity with a heuristic pipeline would require a highly intricate system, involving mesh segmentation, plane fitting and denoising, curve fitting, stitch querying, and more.
In such a large and fragile system, even minor errors could cascade and destabilize the entire pipeline, making it difficult to achieve reliable results.
Therefore, we advocate an approach that directly learns the parsing of \interRep using auto-regressive transformers.

To further investigate a suitable representation of sewing patterns, the latest work~\cite{nakayama2024aipparel} proposes a sewing pattern tokenization scheme to compress the tedious pattern information to structural tokens, and transfer it to a sequence prediction problem.
Following~\cite{nakayama2024aipparel}, we adopt a tokenization scheme for every pattern in GCD:
\begin{equation}
\mathrm{E_g} = \texttt{<SoG>} \, \mathrm{E_p}(P_1, S) \cdots \mathrm{E_p}(P_N, S) \, \texttt{<EoG>}
\end{equation}
Here, $\mathrm{E_g}(P)$ denotes the full sewing pattern sequence composed of $N$ panels. The tokens \texttt{<SoG>} and \texttt{<EoG>} indicate the start and end of the garment, respectively. $\mathrm{E_p}(P_i, S)$ represents the token sequence corresponding to the $i$-th panel $P_i$, along with its associated stitching information $S$.
Each panel is further tokenized into a structured sequence:
\begin{equation}
\mathrm{E_p} = \texttt{<SoP>} \, [\text{name}] \, \texttt{<R>} \, [\text{edges with stitching tags}] \, \texttt{<EoP>}
\end{equation}
The token \texttt{<SoP>} marks the beginning of a panel, followed by a name token identifying the type of the panel, and a pose token \texttt{<R>} encoding its 3D rigid transformation (rotation and translation), and ends with token \texttt{<EoP>}.

Under the above token-based representation, we first sample points $\mathbb{P}_{boxmesh}$ on predicted \interRep $\mathcal{M}$.
Similarly, $\mathbb{P}_{boxmesh}$ is passed to the pretrained shape autoencoder~\cite{zhao2024michelangelo} and an auto-regressive transformer $\mathcal{O}_2$ to predict a pattern token sequence $\mathrm{\tilde{E_g}}$.
Then $\mathrm{\tilde{E_g}}$ is further de-tokenized according to the defined rules to the final sewing pattern $\{\mathcal{\tilde{P}}, \mathcal{\tilde{S}} \}$.
The process can be expressed as:
\begin{equation}
\{\mathcal{\tilde{P}}, \mathcal{\tilde{S}} \} = \mathcal{D}_{seq2pattern}(\mathcal{O}_2(\mathcal{E}_2(\mathbb{P}_{boxmesh}))).
\end{equation}

During training, we follow~\cite{nakayama2024aipparel} to define the total loss as a combination of cross-entropy (CE) on the discrete tokens (such as panel name, edge type, and stitch tag) and $L_2$ loss on the continuous parameters (such as 3D placement, edge vertices, and curve-specific parameters).

%% file: figs/pattern_rep.tex
\begin{figure*}
\centering
    \includegraphics[width=0.9\textwidth]{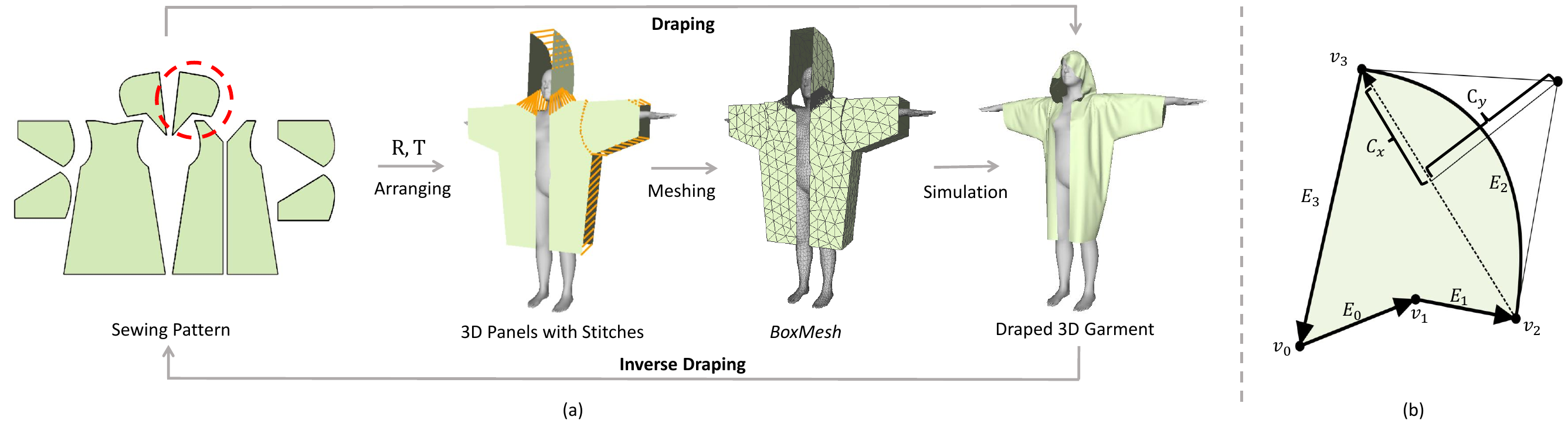}
    \caption{  
        (a) Recovering sewing patterns from a 3D garment mesh is an inverse problem of garment draping. (b) An example of the parameterization of the half hood panel (red circle in (a)). 
    }
    \label{fig:sewing_pattern}
\end{figure*}

%% file: figs/pipeline.tex
\begin{figure*}
\centering
    \includegraphics[width=1.0\textwidth]{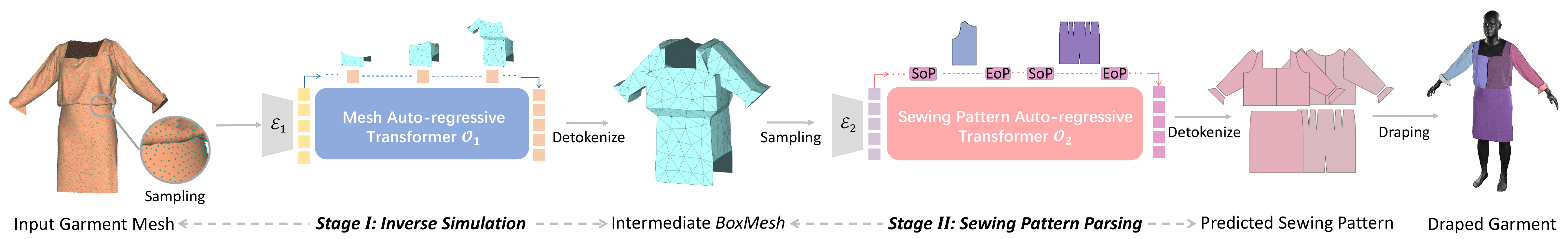}
    \caption{
        Method overview. Given an input garment mesh, our method first predicts an intermediate representation (\interRep) using a geometry-oriented auto-regressive model. A second, semantic-aware model then generates the corresponding sewing pattern, which can be draped onto the body for final garment reconstruction.
        }
    \label{fig:pipeline}
\end{figure*}

%% file: sec/5_experiments.tex
\section{Experiments}
\label{sec:experiments}


\subsection{Datasets}
\label{sec:data}
We follow the most recent works~\cite{nakayama2024aipparel, bian2024chatgarment} of sewing pattern reconstruction and adopt the GCD v2~\cite{GarmentCodeData:2024} for training and evaluation of all modules.
GCD provides two subsets: garments aligned with a default template body, and garments aligned with 5,000 randomly sampled body shapes.
Since both real scanned garments and 3D assets designed for virtual characters typically conform to diverse body shapes, we use the subset aligned with random body shapes. Each sample includes a 3D garment mesh draped over an A-pose body, along with its sewing pattern parameters and a corresponding intermediate \interRep.
GCD contains approximately 132,670 garments across a wide range of categories, where 115,191 garments have augmentation on random body shapes.
We randomly split the dataset into 95\% : 5\% : 5\%, finally resulted in 103,725 for training, and 5733 for validation and testing.

To further assess the generalization capability of our method beyond the synthetic garments in GCD, we evaluate it on a variety of external, real-world garment data.
Unlike GCD, where garments are generated through relatively simple simplified scenarios on template bodies with fixed fabric properties and force settings, real-world garments often exhibit richer details and more irregular deformations.
We collect our own in-the-wild 3D scans of people wearing diverse everyday clothing using a mobile phone (All human subjects provided informed consent for the use of their images in this study).
Additionally, we test our method on publicly available scanned datasets such as THUman2.0~\cite{Zheng2019DeepHuman} and RenderPeople, where garments are first segmented and then retargeted to rest-pose bodies for consistency.
We also construct a test set consisting of an input image and its corresponding 3D garment mesh to to assess the applicability of single-view reconstruction (\cref{subsec: single_view}).

\subsection{Data Pre-processing}
\label{subsec:process}
\textit{\textbf{Pre-processing for GCD.}}
The \interRep provided in GCD is highly detailed, often with dense triangle faces.
To reduce the learning complexity and ensure efficient execution in Stage I, we perform Isotropic Explicit Remeshing to produce more uniform and compact representations.
Specifically, we compress each \interRep to fewer than 1,600 faces, which is sufficient to preserve its detailed shape while remaining within the maximum number of supporting faces reported by~\cite{weng2024scaling}.
For the input of Stage I and Stage II, we uniformly sample 4,096 points from the garment mesh and the predicted \interRep, respectively, as inputs to our modules.

\textit{\textbf{Pre-processing for Cost-effective Real Scan.}}
For real scan data, we first collect multi-view images or a video of volunteers from a local area using a mobile phone.
Then we reconstruct the 3D Gaussians~\cite{kerbl20233d} of the clothed human by~\cite{ye2024gaustudio}.
Note that the data is normalized to a coordinate space ranging from $[-1,1]$, which facilitates consistency across further stages of the pipeline. 
Next, we utilize the SMPL model~\cite{loper2023smpl} to fit the body from the 3D scans. 
The process involves rendering the 3D scan using PyTorch3D~\cite{ravi2020accelerating} from 20 different viewpoints around a full 360$^\circ$ angle.
OpenPose~\cite{cao2019openpose}, through Sapiens~\cite{khirodkar2024sapiens}, is employed to detect the human pose in these rendered views.
Using the detected pose, we estimate the 3D pose and fit the SMPL model to the scan data.

We need to segment the garment part from the full body by methods~\cite{kim2024gala}.
To be specific, the 3D clothed human is rendered into 36 different views. These 2D views are then segmented using~\cite{liu2025grounding} with specific text prompts to identify different clothing areas, such as 'upper cloth', 'lower cloth', and 'wholebody cloth'. 
The 2D segmentation is then re-projected to 3D and used to vote for the segments of the 3D mesh.

To deal with scans from different poses, we simulate the cloth into the A-pose SMPL template body by CounterCraft~\cite{grigorev2024contourcraft}. 
We export the body morphing sequence from the fitted SMPL body of the scan to A-pose first and then obtain the scanned garment worn on the A-pose body after applying the simulation of CounterCraft~\cite{grigorev2024contourcraft}.

\subsection{Implementation}

\textit{\textbf{Training and Inference Details.}}
For Stage I, to prepare training data, garment meshes are first discretized with a voxel resolution of $128 \times 128$.
Then the processed meshes were tokenized using Blocked and Patchified Tokenization (BPT)~\cite{weng2024scaling} to serve as training data.
We use the pre-trained weights from~\cite{weng2024scaling}.
The finetuning of the pretrained model was conducted on 4 Nvidia RTX 6000 Ada GPUs for 20 epochs using about 2 days, with a learning rate of 1e-4 and the Adam optimizer. 
The maximum output sequence length of the model is 10,000, and the training batch size is 32 in total. 

For Stage II, to generate training data, we employed the tokenization scheme in~\cite{nakayama2024aipparel} to convert GCD~\cite{GarmentCodeData:2024} into token sequences and their corresponding continuous variables. Training was conducted on 4 NVIDIA RTX 6000 Ada GPUs, with a batch size of 48 per GPU, a learning rate of 1e-5, and the AdamW optimizer. Initially, the model was trained with the ground-truth \interRep of GCD for 100 epochs using about 2 days to obtain a preliminary weight, followed by finetuning this weight on \interRep predicted by our Stage I module for 20 epochs using about 12 hours.

For inference, it takes about 30s on Stage I, and 3s on Stage II for a single garment. 
The approximate inference time is obtained on a single Nvidia RTX 6000 Ada GPU.

\textit{\textbf{Modification of NeuralTailor.}}
To fairly compare our method with~\cite{korosteleva2022neuraltailor}, we need to modify the original code to accommodate its training on the newly designed data format in GCD~\cite{GarmentCodeData:2024}.
Specifically, we expand the per-panel query embedding from
its default number of 23 to 75 to accommodate all the different panel classes in GCD.
Similarly, we expand the per-edge embedding from 14 to 37. 
Furthermore, because GCD contains cubic Bézier curves and arcs, which the previous dataset~\cite{korosteleva2021generating} does not have, we also extend per-edge parameterization from only using four channels to ten channels (2+4+3+1: two endpoints, four control points, three arc parameters, and one curve type class label).
When the class label is set to 1, the third to sixth channels are the control points of Bézier curves.
If the class label equals -1, the third to sixth channels are left zero, and the remaining channels represent three parameters of arc (radius, large arc flag, sweep flag).
We keep the other settings the same as~\cite{korosteleva2022neuraltailor}.
We train the modified version \textit{NeuralTailor*} on the same training split of our method until its convergence with a learning rate of 0.002 and a batch size of 32 on a single NVIDIA GeForce RTX 3090 GPU for 30 epochs using about 15 hours. 

\textit{\textbf{Body Measurements of SMPL Models.}}
To drape the final output onto various body shapes, we have estimated the underlying body shape, potentially in arbitrary poses, from the scanned clothed human.
Since most existing human-body fitting algorithms are based on the SMPL model~\cite{loper2023smpl}, we adopt SMPL as our canonical body representation as described in~\cref{subsec:process}.
However, the draping system of GCD is build on a custom statistical body model based on CAESAR~\cite{robinette2002civilian}.
It accepts only body measurements as input and does not support direct use of SMPL parameters or meshes. 
Therefore, for each subject, we compute body measurements from their fitted SMPL mesh.
Following the instructions provided in the GCD dataset, we annotate standard landmarks on the SMPL template mesh (based on vertex IDs), and compute each measurement using corresponding geometric definitions, such as angles, geodesic distances, Euclidean distances, or vertical (Y-axis) height differences, according to the detailed documentation available at the official GCD website.
For scans with non-A-pose postures, we convert the estimated SMPL pose into a canonical A-pose configuration (with the upper arms forming a 40° angle with the horizontal plane) before computing measurements, ensuring consistency across different inputs.

\subsection{Metrics}
\label{sec:metrics}
In this subsection, we summarize the standardized metrics used for fair and consistent quantitative evaluation and comparison.

\textit{\textbf{Metrics on BoxMesh Prediction.}}
To assess the quality of the predicted \interRep against the ground-truth \interRep, we uniformly sample 10,000 points from both the ground-truth meshes and the generated meshes and compute similar evaluation metrics as~\cite{weng2024scaling} on the evaluation dataset of GCD:
\begin{itemize}[leftmargin=*]
  \item \textbf{Chamfer Distance (CD)}: Evaluates the similarity between two point clouds by averaging the distances from each point to its nearest neighbor in the other set.
  \item \textbf{Hausdorff Distance (HD)}: Assesses the maximum deviation between two point sets by taking the greatest of all nearest-neighbor distances across both sets.
\end{itemize}

\textit{\textbf{Metrics on Sewing Pattern Prediction.}}
To evaluate the quality of sewing pattern predictions, we adopt a set of metrics similar to those used in~\cite{korosteleva2022neuraltailor, liu2023sewformer} and our new metrics:

\begin{itemize}[leftmargin=*]
\item \textbf{Panel $\mathbf{L_2}$}: Measures the geometric accuracy of each panel by computing the vertex-wise ${L_2}$-distance between the predicted and ground-truth 2D panel shapes.
\item \textbf{Transl. $\mathbf{L_2}$ \& Rot. $\mathbf{L_2}$}: Evaluate the accuracy of the predicted 3D placement of panels in terms of their translations and rotations compared to the ground truth.
\item \textbf{\#Panels Acc. \& \#Edges Acc.}: Quantify the correctness of the predicted garment structure by comparing the number (denoted by ``\#") of panels and edges with those in the ground truth.
\item \textbf{Stitch Prec.}: Assesses how accurately the model predicts stitch connections between panels.
\item \textbf{Panel IoU}: Computes the Intersection over Union (IoU) between predicted and ground-truth 2D panel masks, offering a region-based evaluation of panel shape quality.
\item \textbf{CD \interRep}: Reflects the overall quality sewing pattern in 3D by calculating the Chamfer Distance of the \interRep from reconstructed pattern and ground-truth \interRep.
\end{itemize}
While Panel ${L_2}$~\cite{korosteleva2022neuraltailor} is widely used, it only measures vertex-wise distances and is sensitive to small geometric deviations, especially in synthetic datasets with rigid panel templates. Yet, sewing panels are defined by their overall shape and area, not just point locations. To better capture this and provide a more comprehensive evaluation, we introduce Panel IoU, which compares 2D panel regions directly.
For example, if a straight edge is predicted as two slightly angled segments, the ${L_2}$ error may be large, but the panel region remains intact, resulting in a stable and accurate IoU.
Besides, we also propose to use CD \interRep to measure the overall quality of the reconstructed sewing pattern in 3D space, since \interRep contains all the information in the level of panel shape fidelity, 3D placement accuracy, and stitch correctness. Note that this \interRep is constructed from the final sewing pattern prediction and is NOT the \interRep predicted in Stage I.

To fairly compare different methods, it is necessary to consider all of the above metrics on predicted sewing patterns, as any individual metric alone may be noisy and lead to a biased evaluation.

\subsection{Evaluation}
\input{figs/tab_eval_stage1}
\input{figs/eval_stage1}

\input{figs/tab_eval_stage2}
\input{figs/tab_compare_nt}

\textit{\textbf{Evaluation on Inverse Simulation.}}   
We first evaluate our Stage I, i.e., the \stageI module, both quantitatively (\cref{tab:eval_stage1}) and qualitatively (\cref{fig:eval_stage1}).
We compare the Compressive Tokenization used in our method with a Direct Tokenization strategy that encodes mesh geometry via raw vertices and faces, similar to~\cite{chen2024meshanything}.
Thanks to its compact and structured representation, Compressive Tokenization improves the learning of our 'inverse simulation' task.
As shown in the table, our method achieves lower Chamfer Distance (CD) and Hausdorff Distance (HD) on the GCD testing set, indicating better geometric reconstruction of \interRep.
Qualitatively, the predicted \interRep generated with Compressive Tokenization more closely matches the ground truth and preserves the sharp features.
In contrast, using Direct Tokenization often results in noticeable structural inconsistencies, especially in fine regions with sharp features (red boxes) such as hems, collars and cuffs.

\input{figs/eval_stage2}

\textit{\textbf{Evaluation on Sewing Pattern Parsing.}} We evaluate our Stage II, i.e., the \stageII module, with both quantitative metrics (\cref{tab:eval_stage2}) and qualitative results (\cref{fig:eval_stage2}).
We first train the module using ground-truth \interRep as input and refer to this model \textit{Ours-GT}.
Then we test it using the ground-truth \interRep as input, whose results are denoted as \textit{Ours-GT} in~\cref{tab:eval_stage2}.
To assess the influence of upstream prediction errors from Stage I, we test this same model with predicted \interRep instead, denoted as \textit{Ours-GT*} (* denotes that the input BoxMesh is from Stage I during testing).
As expected, performance drops in \textit{Ours-GT*} obviously, indicating that the inaccuracies in predicted \interRep degrade the final pattern reconstruction.

To mitigate this issue, we fine-tune the above model by training it on predicted \interRep (in the training set) from Stage I.
The fine-tuned model is denoted as \textit{Ours-Pred}, and its predicted-input testing variant is denoted as \textit{Ours-Pred*}.
As shown in~\cref{tab:eval_stage2}, \textit{Ours-Pred*} achieves noticeable improvements over \textit{Ours-GT*}, showing that fine-tuning helps the model adapt to imperfections in the input.
This improvement is also clearly visible in the qualitative comparisons in~\cref{fig:eval_stage2}.
We observe that the input \interRep in column (d) from Stage I usually contains extra noise, which may be out of distribution from the ground-truth. 
If the network only learn from purely clean data, it could be sensitive to the quality of the input \interRep. 
As a result, it tends to predict redundant structures and irregular panel shapes.
On the other hand, fine-tuning makes the module more robust to the noise, which may be especially useful when encountering real-world data.

\subsection{Comparisons}
We compare our method with the most relevant prior work, NeuralTailor~\cite{korosteleva2022neuraltailor}.
For a fair comparison, we retrain NeuralTailor on GCD with the same training split as our method (data with random body shapes) with slight modifications (NeuralTailor*) to its original architecture.

\textit{\textbf{Quantitative Comparisons.}}   
For quantitative evaluation, we compare our method against \textit{NeuralTailor*} on the GCD testing set with random body shapes.
We compute the metrics described in \cref{sec:metrics} for the predicted sewing patterns. As shown in~\cref{tab:compare_nt}, our method consistently outperforms \textit{NeuralTailor*} across all metrics.

\textit{\textbf{Qualitative Comparisons.}}
As described in~\cref{sec:data}, we compare our method with \textit{NeuralTailor*} on garments outside the GCD distribution. 
These include garments from our own cost-effective real scans and existing datasets.
Before being input into our pipeline, the 3D scans are processed into raw 3D garments and then deformed to be worn on the rest pose, as the process described in~\cref{subsec:process}.
As shown in~\cref{fig:compare_nt}, our method produces reasonable and structurally faithful sewing patterns across diverse input sources (the first two columns are from RenderPeople and the last two are from our own scans), whereas \textit{NeuralTailor*} struggles to generalize to real data, particularly in stitch prediction, which often results in failed draping. We have also listed the \interRep constructed from the predicted sewing pattern for comprehensive comparison. 
More results for sewing pattern recovery from real scans of our method can be found in~\cref{fig:more_results}.

\input{figs/compare_nt}

\input{figs/tab_ablation}

\input{figs/ablation_baseline}
\input{figs/ablation_default}
\input{figs/ablation_points}
\input{figs/ablation_extraOutput}

\subsection{Ablation Study} 
To better understand our method, we conduct an ablation study with the following five configurations:

\begin{itemize}[leftmargin=*]
    \item \textbf{\textit{Baseline}:} We directly train the Stage II network to map sampled points from the garment mesh to the sewing pattern, without using the intermediate \interRep representation, which removes our core design.
    \item \textbf{\textit{Default\_Body}:} We train our full pipeline on the subset of GCD where garments are draped over the same standard template body shape (default), instead of using the diverse random body shapes.
    \item \textbf{\textit{Inter\_Points}:} We directly generate point clouds of \interRep in Stage I as the input of Stage II, instead of first generating an intermediate mesh and then sampling points from it. Stage II is also fine-tuned on the predicted results of Stage I.
    \item \textbf{\textit{Extra\_Output}:} Following the architecture of \textbf{\textit{Baseline}}, \interRep is not used as an intermediate representation, but as an additional final output alongside the sewing patterns. 
    \item \textbf{\textit{Full}:} Our full two-stage method, including fine-tuning on predicted \interRep and training on garments with random body shapes.
\end{itemize}

As shown in Table~\ref{tab:ablation}, our full method outperforms all ablated variants overall, confirming the effectiveness of our design.
To better support the advantages of our full model, we also provide visual results in~\cref{fig:ablation_baseline}, \cref{fig:ablation_default}, \cref{fig:ablation_points} and~\cref{fig:ablation_extraOutput}.

Comparing \textit{Baseline} and \textit{Full}, we observe a significant improvement when using the intermediate \interRep representation. 
From~\cref{fig:ablation_baseline}, we observe that \textit{Baseline} can only predict a rough shape from the input garment mesh while our full model can capture accurate features such as the hem of a strapless top or a fishtail dress.  
This demonstrates the benefit of decomposing the complex mapping from 3D garment mesh to sewing pattern into two easier sub-tasks: inverse simulation and sewing pattern parsing. The \interRep serves as a structured intermediate representation that simplifies the learning process and enhances the model’s ability to recover more accurate garment structure.

\input{figs/comparison_mm}

The comparison between \textit{Default\_Body} and \textit{Full} further highlights the importance of considering diverse body shapes during training. We would like to claim that we are the first to use the body-shape-diverse training data of GCD and the first to verify its effectiveness.
The model trained solely on garments draped over the default body shape performs worse, likely because it fails to capture how body shape variations influence panel scale and fabric behavior during simulation. 
In \cref{fig:ablation_default}, the model of \textit{Default\_Body} fails to reflect the actual size of an input garment.
In contrast, our setting better reflects real-world variability and leads to improved generalization.

In Stage I, an alternative approach is to directly generate point clouds in the shape of \interRep. However, point clouds are inherently irregular and do not guarantee continuity between neighboring points. Moreover, existing point generation methods do not readily support domain transfer between 3D garments and \interRep, making this approach particularly challenging.
We experiment with point diffusion~\cite{wu2024ipod}, using sampled points from the 3D garment as input to directly predict points on \interRep. However, as shown in~\cref{fig:ablation_points}, this approach tends to produce results with artifacts such as holes, floating points, blurred boundaries, and incomplete structures, which undermine the effectiveness of the intermediate representation.
In contrast, mesh generation via auto-regressive modeling produces \interRep with structured surfaces composed of planar panels and smooth stitching. Sampling from such meshes yields more uniform point distributions with better geometric quality, which makes Stage II easier.

We also explore another design in which \interRep is treated as an additional final output alongside the sewing patterns, rather than as an intermediate representation. 
Specifically, we construct two output sequences simultaneously, corresponding to the sewing pattern parameters and mesh faces, respectively.
However, these two outputs lack explicit correspondence, making the joint learning difficult. 
As a result, it fails to produce accurate predictions. 
As shown in~\cref{fig:ablation_extraOutput}, the generated results exhibit artifacts similar to those observed in the \textit{Baseline}.
These observations suggest that using \interRep as an intermediate representation is essential in our approach.

Overall, these results validate the necessity and effectiveness of our design choices, including the two-stage architecture, the use of body-shape-diverse training data, and how to utilize the intermediate representation.

\input{figs/tab_multi_modal}

\subsection{Application on Single-view Reconstruction}
\label{subsec: single_view}
Recent works~\cite{nakayama2024aipparel,bian2024chatgarment} based on the GCD dataset typically take single-view images (or a combination with textual description) as input, making them more user-friendly.
Although our method is originally designed to reconstruct sewing patterns from garment meshes, it can also be extended to the single-view setting, for example when only online images are available.
In this case, given a single-view image, we first reconstruct a 3D clothed human mesh using 3D foundation models such as Hunyuan3D~\cite{zhao2025hunyuan3d}, then segment the garments from the reconstructed mesh (as described in~\cref{subsec:process}), and finally apply our pipeline.
We refer this application as Ours-SVR.

We compare Ours-SVR with the state-of-the-art methods AIpparel~\cite{nakayama2024aipparel} and ChatGarment~\cite{bian2024chatgarment} on a photorealistic synthetic image test with pseudo-ground-truth meshes.
To construct the test set, we first generate 40 high-quality full-body images of clothed humans in rest pose with diverse garment styles using GPT-4o. We then reconstruct 3D clothed human meshes with Hunyuan3D and segment the garment meshes to serve as ground truth.
Since both AIpparel and ChatGarment are trained only on data of default body shape, draping their predicted sewing patterns onto the predicted bodies sometimes fails. Therefore, for consistency, we drape the results of these two methods onto the default template body.
For evaluation, we follow ChatGarment~\cite{bian2024chatgarment} and report Chamfer Distance and $F_1$ scores under two distance thresholds (in centimeters) between the recovered meshes (obtained by generating sewing patterns and re-draping onto the A-pose body) and the ground-truth meshes.

As shown in~\cref{fig:compare_mm} and~\cref{tab:compare_single_view}, Ours-SVR achieves comparable performance, although it is not specially designed for single-view reconstruction. 
Our recovered meshes even better reflect the underlying sewing pattern design when only single-view cues are available. 
ChatGarment~\cite{bian2024chatgarment} demonstrates a high simulation success rate and produces visually reasonable results, but sometimes fails to capture key garment details such as asymmetric sleeve designs or whether a jacket is open or closed.
AIpparel~\cite{nakayama2024aipparel} generalizes poorly due to the limitation of its underlying synthetic training data. 
In addition, AIpparel can only process images where the person is wearing a single garment. 
Since people in real life rarely wear only a top or only a bottom, in~\cref{tab:compare_single_view} we evaluate AIpparel only on 18 whole-body garments, such as dresses and jumpsuits.
For fairness, we also report the results of Ours-SVR and ChatGarment on this subset, which are marked with an asterisk (*).
More results of our method for sewing pattern recovery from single images can be found in~\cref{fig:more_results}.

\subsection{Limitation and Discussion}  
Our method may fail in certain scenarios. For instance, as shown in~\cref{fig:failure} (left), garments with extreme length, such as an Arabic Kandora with an occluded lower region, often lead to noisy or incomplete geometry in the scanned mesh. As our pipeline does not include manual mesh cleanup, such artifacts can propagate through the reconstruction process, resulting in unreliable sewing patterns, particularly in stitching relationships, and may ultimately cause draping failures during simulation. The example in~\cref{fig:failure} (right) further highlights a limitation of the underlying synthetic dataset, which lacks coverage of certain garment types, such as pleated skirts. Extending sewing pattern datasets to better capture the diversity of real-world garments remains an important direction for future work.

\input{figs/failure_cases}

\input{figs/more_results}

%% file: figs/tab_eval_stage1.tex
\begin{table}
\caption{Evaluation on Garment-to-BoxMesh on GCD testing set. }
 \begin{center}
  \begin{tabular}{l|c|c}
   \hline
    {Methods}& CD $\downarrow$ & HD $\downarrow$ \\
    \hline
    {\textit{Direct Tokenization}} & 6.46 & 12.66 \\
    {\textit{\textbf{Compressive Tokenization}}} & 3.97 & 9.19  \\

   \hline
  \end{tabular}
 \label{tab:eval_stage1}
 \end{center}
\end{table} 

%% file: figs/eval_stage1.tex
\begin{figure}
\centering
    \includegraphics[width=0.48\textwidth]{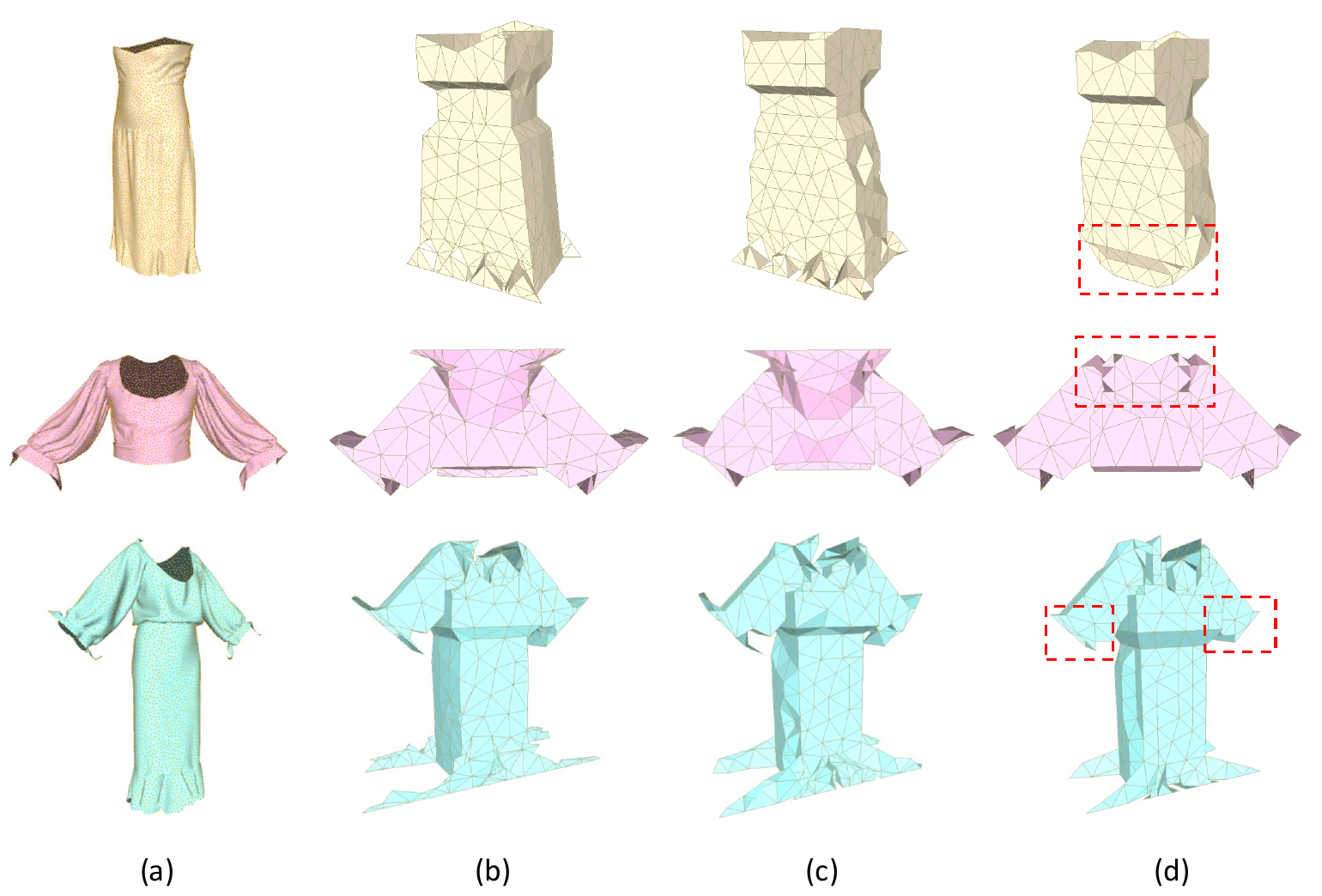}
    \caption{  
        Evaluation on Stage I. From left to right: (a) the 3D garment with sampled points, (b)-(d) \interRep of ground truth, predicted by Compressive Tokenization (our results), and Direct Tokenization, respectively.
    }
    \label{fig:eval_stage1}
\end{figure}

%% file: figs/tab_eval_stage2.tex
\begin{table*}[!h]
\caption{Evaluation on BoxMesh-to-Pattern on GCD testing set. * denotes that the input BoxMesh is from Stage I during testing. \# denotes ``the number of".}
 \begin{center}
  \begin{tabular}{l|c|c|c|c|c|c|c|c}
   \hline
    {Methods}& Panel IoU $\uparrow$ & Panel $L_2$ $\downarrow$ & Rot. $L_2$ $\downarrow$ & Transl. $L_2$ $\downarrow$ & \#Panels Acc. $\uparrow$ & \#Edges Acc. $\uparrow$ & Stitch Prec. $\uparrow$ & CD \interRep $\downarrow$\\
    \hline
    {\textit{Ours-GT}} & 73.23\% & 7.46 & 0.0081 & 4.27 & 67.32\% & 71.85\%  & 69.94\% & 3.64\\
    \hline
    {\textit{Ours-GT*}} & 64.36\% & 11.14 & 0.022 & 10.66 & 34.05\%  & 62.83\%  & 51.56\% & 6.09\\
    {\textbf{\textit{Ours-Pred*}}} & 71.42\% & 9.00 & 0.010 & 5.55 & 57.78\%  & 69.40\%  & 63.47\% & 5.35\\

   \hline
  \end{tabular}
 \label{tab:eval_stage2}
 \end{center}
\end{table*} 

%% file: figs/tab_compare_nt.tex
\begin{table*}[!h]
\caption{Quantitative comparisons with NeuralTailor*~\cite{korosteleva2022neuraltailor}.}
 \begin{center}
  \begin{tabular}{l|c|c|c|c|c|c|c|c}
   \hline
    {Methods}& Panel IoU $\uparrow$ & Panel $L_2$ $\downarrow$ & Rot. $L_2$ $\downarrow$ & Transl. $L_2$ $\downarrow$ & \#Panels Acc. $\uparrow$ & \#Edges Acc. $\uparrow$ & Stitch Prec. $\uparrow$ & CD \interRep $\downarrow$\\
    \hline
    \textit{NeuralTailor*} & 51.88\% & 22.96 & 0.381 & 25.35 & 17.68\%  & 23.09\%  & 18.69\% & 15.46\\
    \textbf{\textit{Ours}} & 71.42\% & 9.00 & 0.010 & 5.55 & 57.78\%  & 69.40\%  & 63.47\% & 5.35\\

   \hline
  \end{tabular}
 \label{tab:compare_nt}
 \end{center}
\end{table*} 

%% file: figs/eval_stage2.tex
\begin{figure}
\centering
    \includegraphics[width=0.48\textwidth]{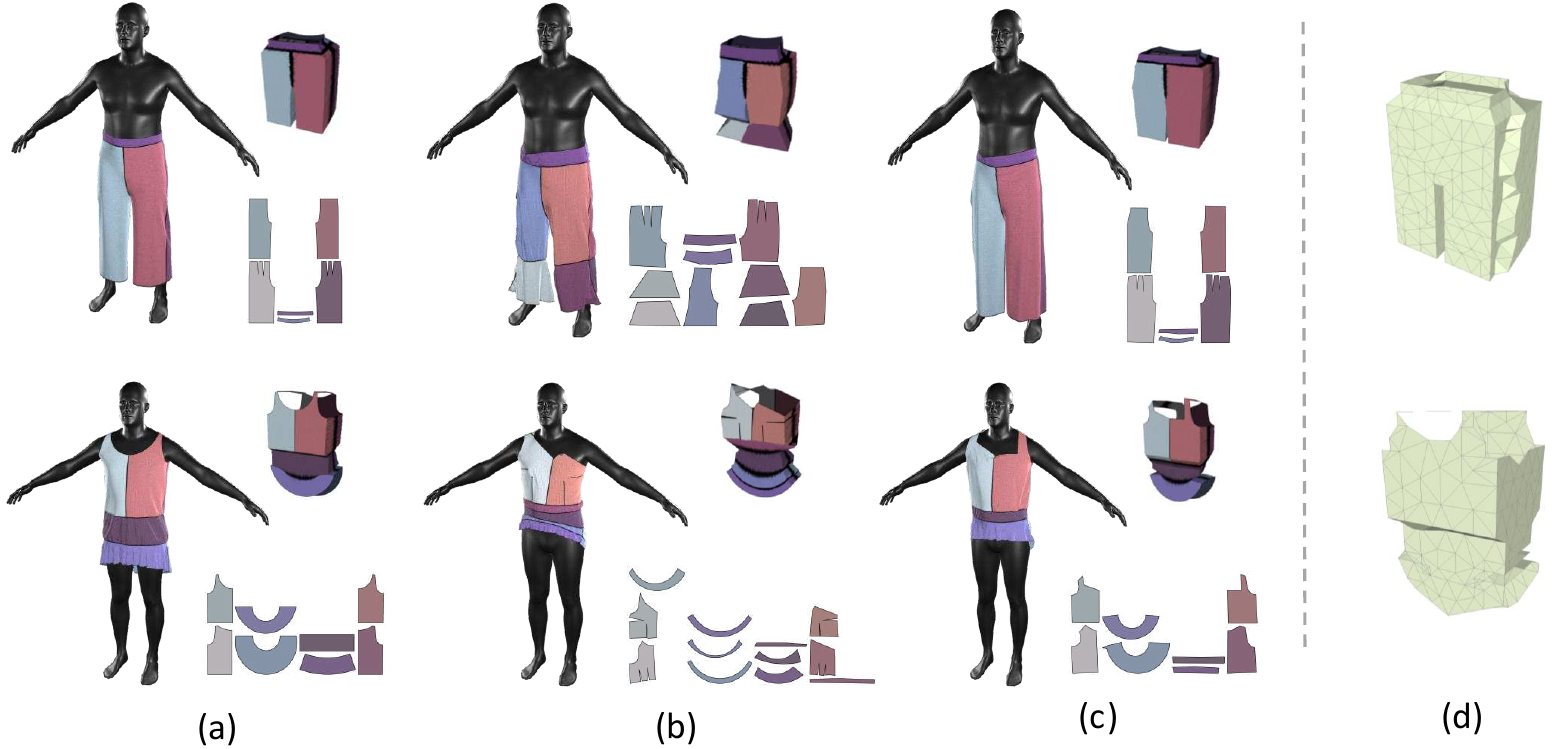}
    \caption{  
        Evaluation on Stage II. From left to right: (a)-(c) the draped 3D garment with sewing pattern displayed in 2D and the corresponding \interRep of ground truth, \textit{Ours-GT*} and \textit{Ours-Pred*}, respectively; (d) the input \interRep predicted from Stage I.
    }
    \label{fig:eval_stage2}
\end{figure}

%% file: figs/compare_nt.tex
\begin{figure*}
\centering
\includegraphics[width=0.9\linewidth]{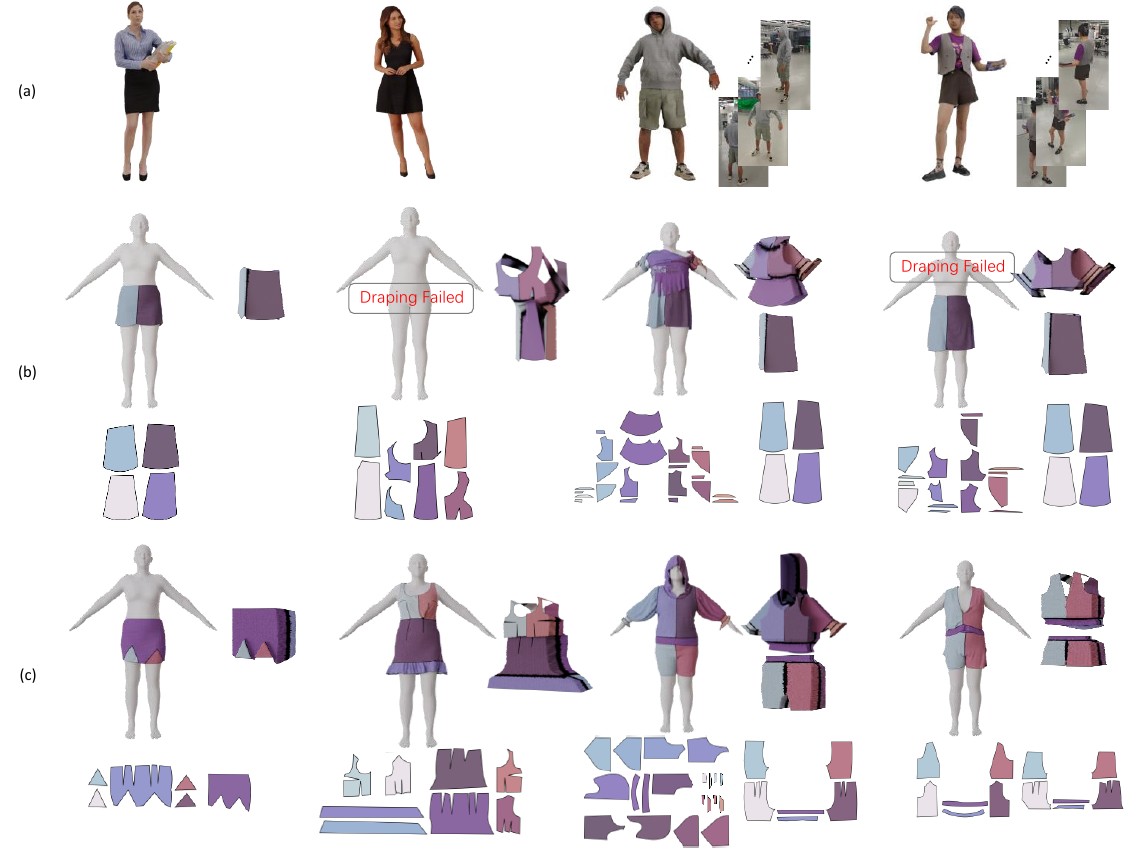}
    \caption{Qualitative comparisons with NeuralTailor~\cite{korosteleva2022neuraltailor} on real-scan data. First two columns are results from RenderPeople, and the rest two columns are from in-the-wild data collected by ourselves. From top to bottom: (a) the input 3D scan, (b)-(c) the simulated garment with corresponding final \interRep and 2D sewing pattern of \textit{NeuralTailor*} and our method, respectively.}

\label{fig:compare_nt}
\end{figure*}

%% file: figs/tab_ablation.tex
\begin{table*}[!h]
\caption{Quantitative comparisons for ablation studies.}
 \begin{center}
  \begin{tabular}{l|c|c|c|c|c|c|c|c}
   \hline
    {Methods}& Panel IoU $\uparrow$ & Panel $L_2$ $\downarrow$ & Rot. $L_2$ $\downarrow$ & Transl. $L_2$ $\downarrow$ & \#Panels Acc. $\uparrow$ & \#Edges Acc. $\uparrow$ & Stitch Prec. $\uparrow$ & CD \interRep $\downarrow$\\
    \hline
    \textit{Baseline} & 68.70\% & 10.26 & 0.015 & 6.37 & 44.44\%  & 65.68\%  & 57.66\% & 5.13\\
   \textit{Default\_Body} & 58.82\% & 14.06 & 0.025 & 10.73 & 37.06\%  & 57.56\%  & 45.13\% & 8.42\\
   \textit{Inter\_Points} & 69.89\% & 9.91 & 0.012 & 6.01 & 50.67\%  & 66.57\%  & 60.03\% & 5.46\\
   \textit{Extra\_Output} & 66.11\% & 11.14 & 0.014 & 6.55 & 47.63\%  & 64.98\%  & 53.80\% & 5.81\\
    \textbf{\textit{Ours}} & 71.42\% & 9.00 & 0.010 & 5.55 & 57.78\%  & 69.40\%  & 63.47\% & 5.35\\

   \hline
  \end{tabular}
 \label{tab:ablation}
 \end{center}
\end{table*} 

%% file: figs/ablation_baseline.tex
\begin{figure}[!t]
    \centering
    \includegraphics[width=0.48\textwidth]{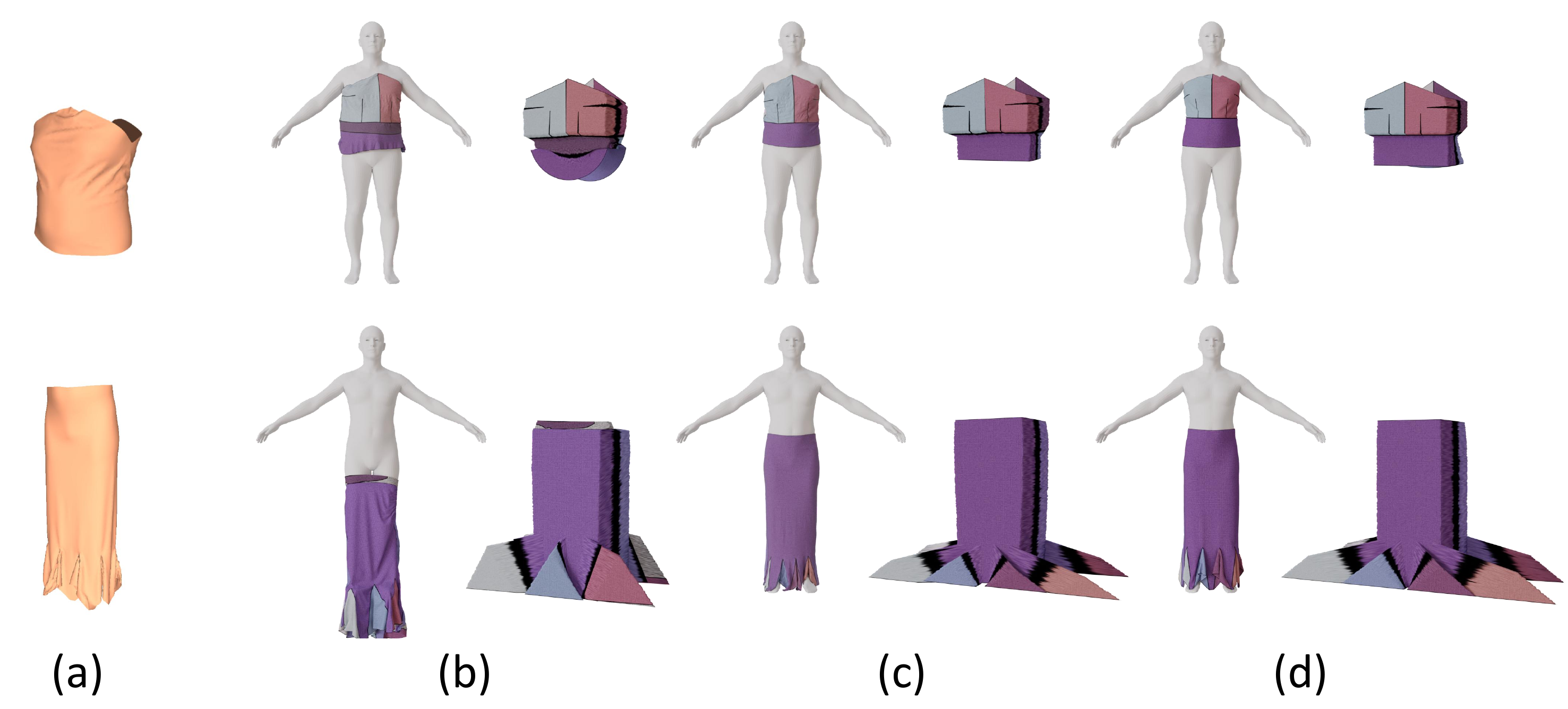}
    \caption{  
        Ablation study with \textit{Baseline}. From left to right: (a) the input garment mesh, (b)-(d) the draped 3D garment and corresponding \interRep from \textit{Baseline},  \textit{Full} and ground truth, respectively.
    }
    \label{fig:ablation_baseline}
\end{figure}

%% file: figs/ablation_default.tex
\begin{figure}[!t]
    \centering
    \includegraphics[width=0.48\textwidth]{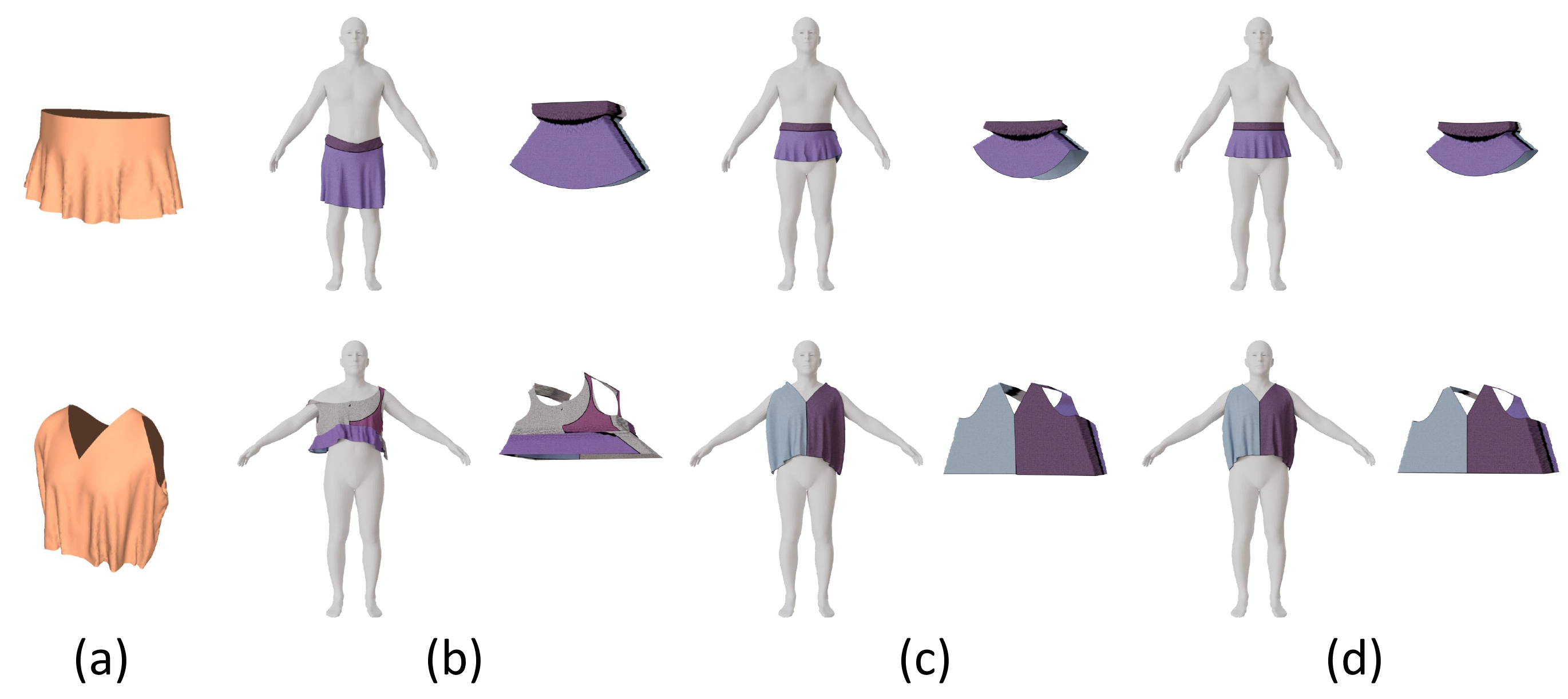}
    \caption{  
        Ablation study with \textit{Default\_Body}. From left to right: (a) the input garment mesh, (b)-(d) the draped 3D garment and corresponding \interRep from \textit{Default\_Body}, \textit{Full} and ground truth, respectively.  
    }
    \label{fig:ablation_default}
\end{figure}

%% file: figs/ablation_points.tex
\begin{figure}[!t]
    \centering
    \includegraphics[width=0.48\textwidth]{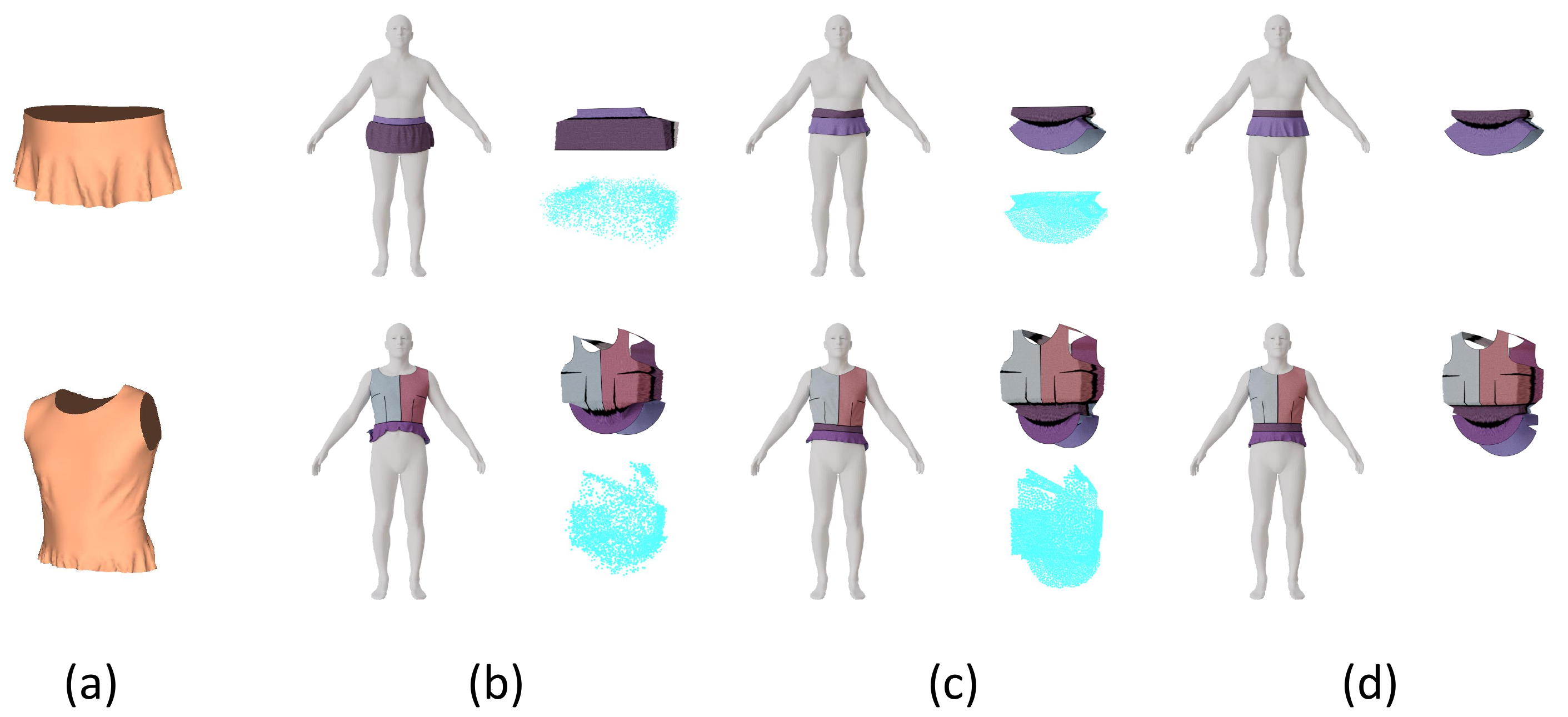}
    \caption{  
        Ablation study with \textit{Inter\_Points}. From left to right: (a) the input garment mesh, (b)-(c) the draped 3D garment, the corresponding \interRep, and the input points of Stage II from \textit{Inter\_Points} and \textit{Full}, (d) ground truth.} 
    \label{fig:ablation_points}
\end{figure}

%% file: figs/ablation_extraOutput.tex
\begin{figure}[!t]
    \centering
    \includegraphics[width=0.48\textwidth]{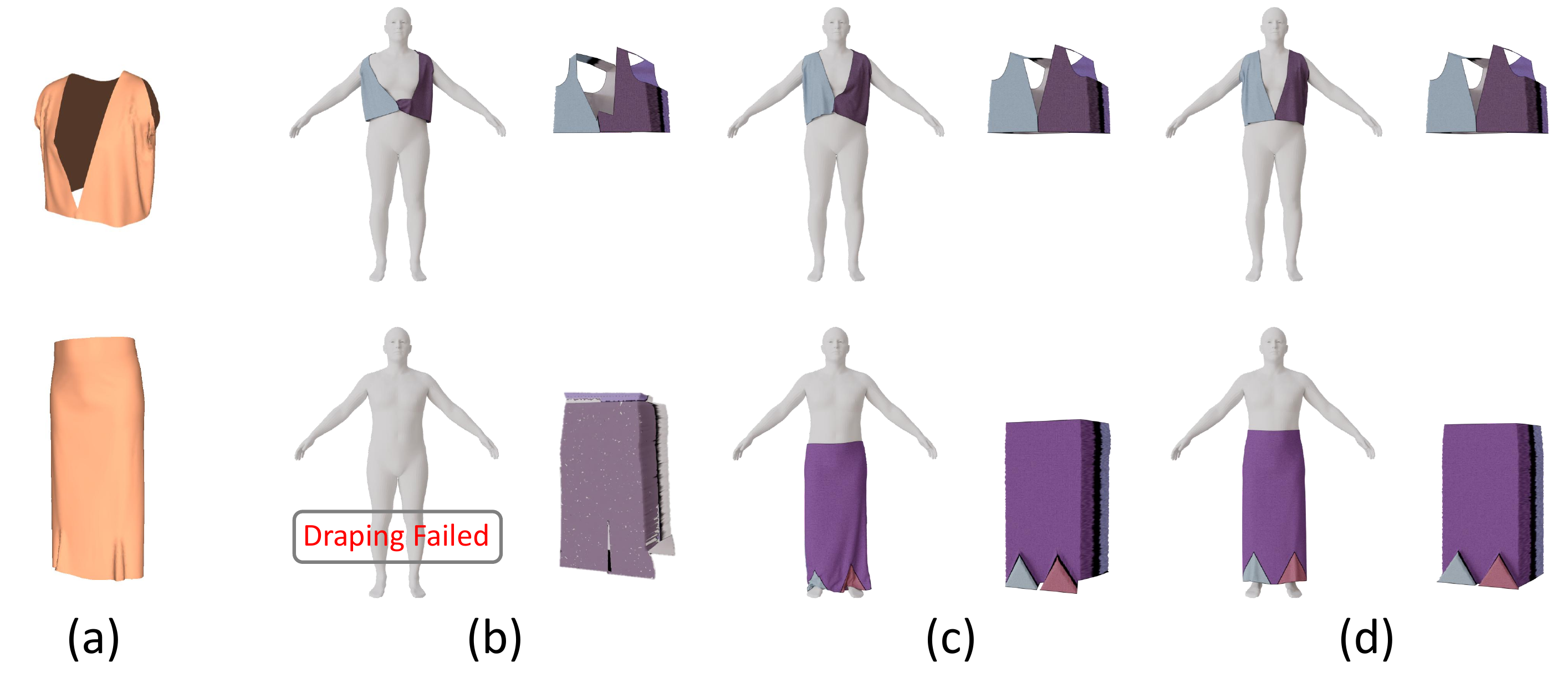}
    \caption{  
        Ablation study with \textit{Extra\_Output}. From left to right: (a) the input garment mesh, (b)-(d) the draped 3D garment and corresponding \interRep from \textit{Extra\_Output},  \textit{Full} and ground truth, respectively.
    }
    \label{fig:ablation_extraOutput}
\end{figure}

%% file: figs/comparison_mm.tex
\begin{figure*}
\centering
    \includegraphics[width=0.95\textwidth]{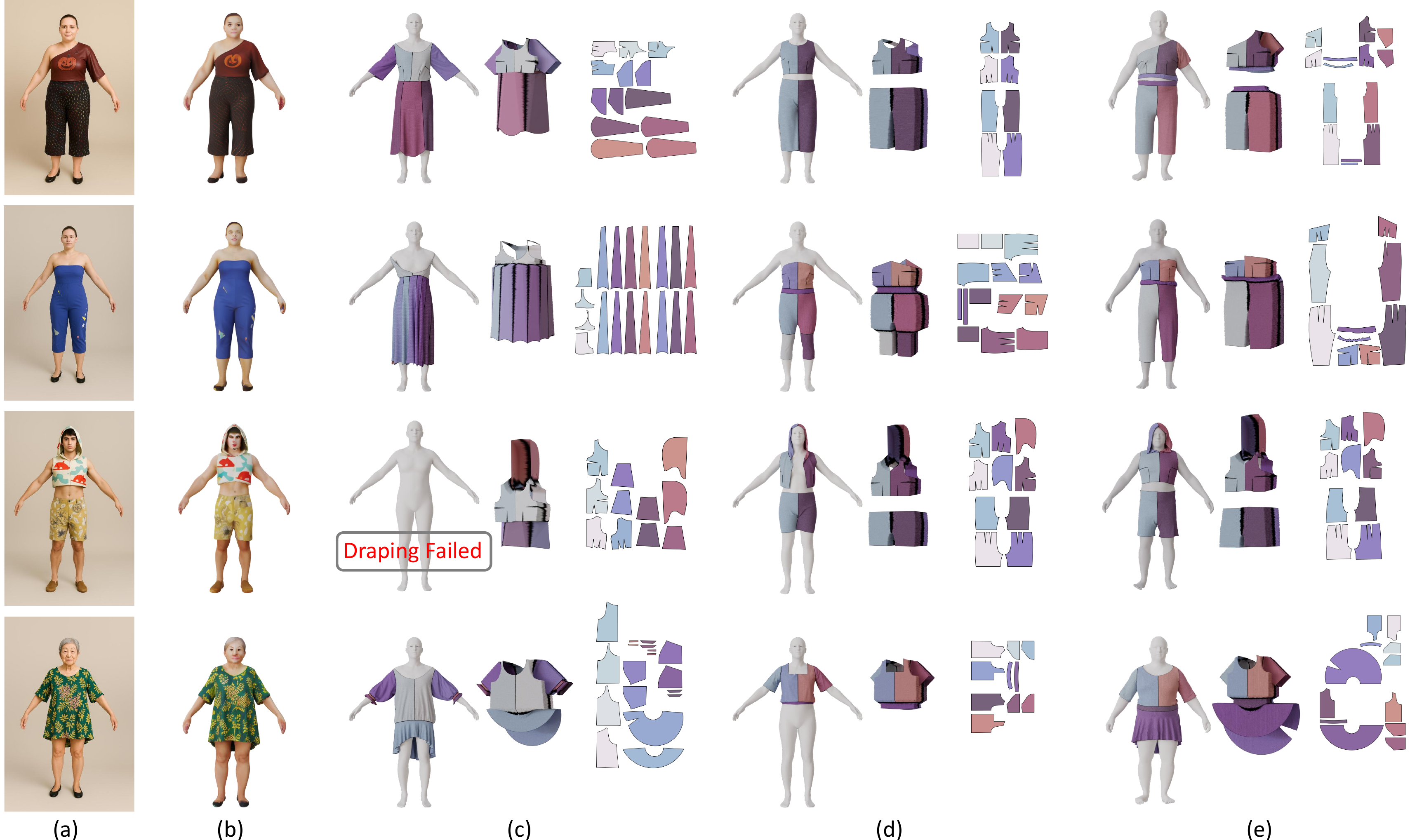}
    \caption{  
        Qualitative comparisons on the application of single-view reconstruction. From left to right: (a) the input image, (b) the generated 3D clothed human mesh, (c)-(d) the draping result with corresponding \interRep and 2D sewing pattern of AIpparel~\cite{nakayama2024aipparel}, ChatGarment~\cite{bian2024chatgarment}, and ours-SVR, respectively.
    }
    \label{fig:compare_mm}
\end{figure*}

%% file: figs/tab_multi_modal.tex
\begin{table}
\caption{Qualitative comparisons with~\cite{bian2024chatgarment}. Metrics are calculated on the intersection of successfully draped samples of all methods.
}
 \begin{center}
  \begin{tabular}{l|cccc}
   \hline
    {Methods}& CD $\downarrow$ & $F_1$\text{@}1cm $\uparrow$ & $F_1$\text{@}3cm $\uparrow$ & Failure Rate $\downarrow$ \\
    \hline
    {\textit{AIpparel}} & 6.09 & 0.16 & 0.62 & 61.1\%  \\
    {\textit{ChatGarment*}} &  6.77 & 0.17 & 0.64 & 0  \\
    {\textit{\textbf{Ours-SVR*}}} & 6.32 & 0.19 & 0.70 & 0 \\
    \hline
    {\textit{ChatGarment}} & 6.58 & 0.19 & 0.63 & 0  \\
    {\textit{\textbf{Ours-SVR}}} & 5.59 & 0.25 & 0.74 & 0 \\
    
   \hline
  \end{tabular}
 \label{tab:compare_single_view}
 \end{center}
\end{table} 

%% file: figs/failure_cases.tex
\begin{figure}[htbp]
    \centering
    \includegraphics[width=0.48\textwidth]{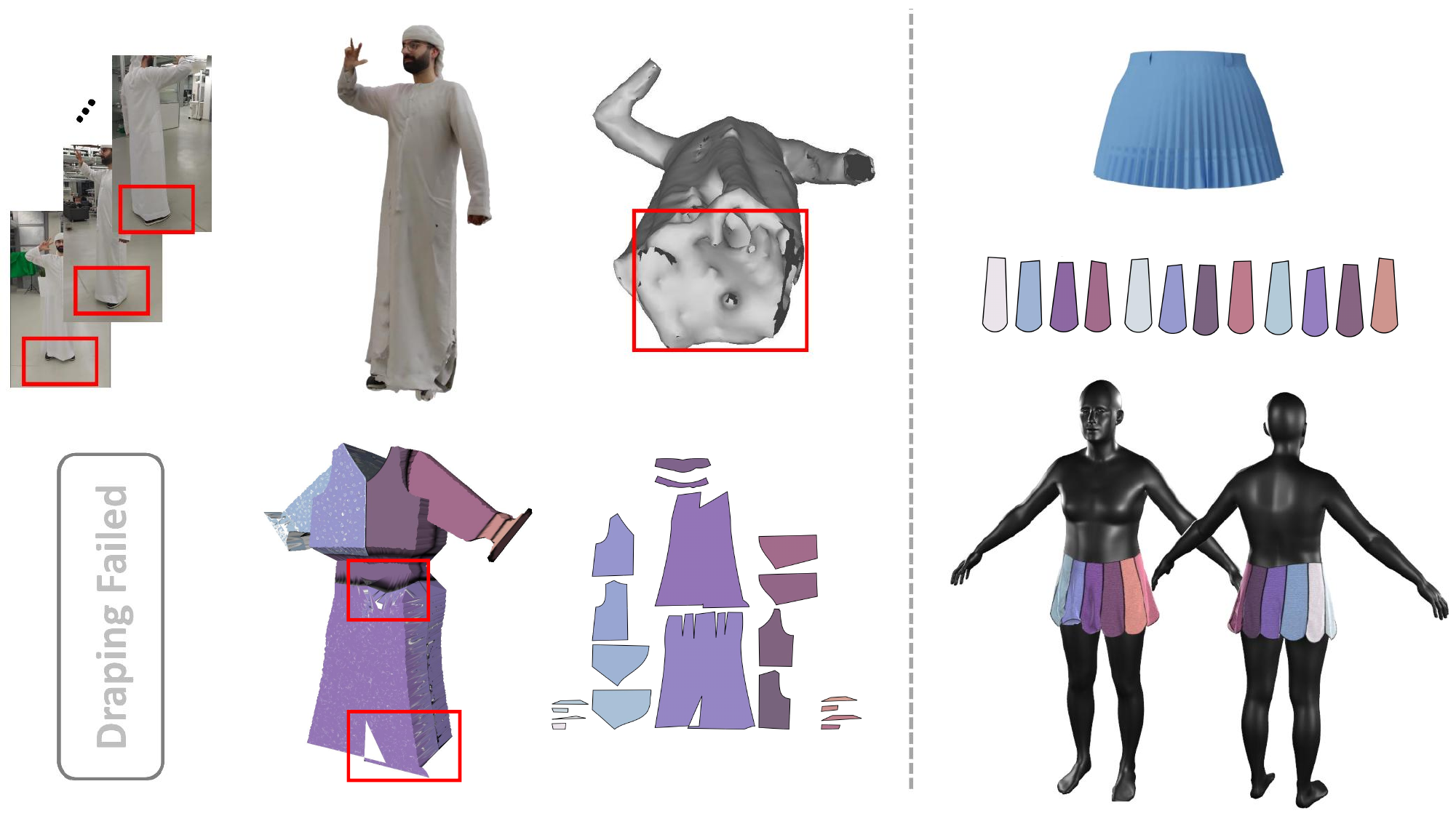}
    \caption{  
        Failure cases. The left example shows errors caused by noise in the scanned garment mesh, while the right example highlights limitations in the design of the current synthetic data.
    }
    \label{fig:failure}
\end{figure}

%% file: figs/more_results.tex
\begin{figure}
\centering
    \includegraphics[width=0.48\textwidth]{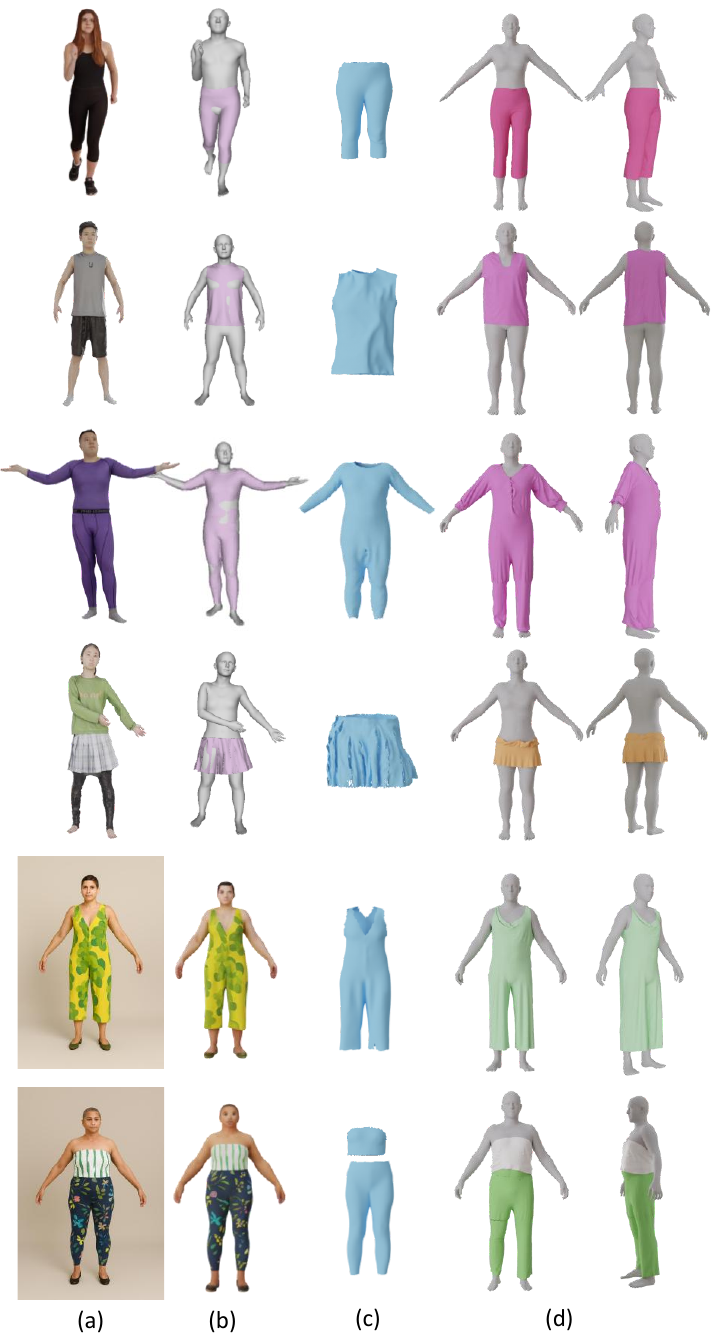}
    \caption{More results of our method. From left to right: (a) input 3D scan (the first row is from RenderPeople and the second to fourth rows are from THuman) or single-view image (last two rows), (b) segmented 3D garment with fitted body or reconstructed 3D clothed human from single view, (c) 3D garment deformed into A-pose, (d) our result after simulation. 
    }
    \label{fig:more_results}
\end{figure}

%% file: sec/6_conclusion.tex
\section{Conclusion}

We presented a two-stage framework for recovering sewing patterns from 3D garment meshes, introducing an intermediate \interRep representation to simplify the complex mapping from deformed garments to structured patterns. Our method demonstrates strong performance on the challenging dataset GCD with diverse body shapes and generalizes to real scans from well-established datasets, cost-effective multi-view scans with cellphones, and even single-view images. Through extensive experiments and ablation studies, we validate the effectiveness of each design component, including the use of intermediate representations, diverse training data, and fine-tuning strategies.